\documentclass[journal]{IEEEtran}
\usepackage{cite}

\usepackage{times}
\usepackage{epsfig}
\usepackage{graphicx}
\usepackage{amsmath}
\usepackage{amssymb}
\usepackage{balance}
\usepackage{url}

\usepackage[pagebackref=true,breaklinks=true,letterpaper=true,colorlinks,bookmarks=false]{hyperref}

\ifCLASSINFOpdf
\else
\fi

\newcommand{\etal}{\emph{et al.}}
\newcommand{\eg}{\emph{e.g.}}
\newcommand{\ie}{\emph{i.e.}}

\begin{document}
%
\title{SG-One: Similarity Guidance Network for One-Shot Semantic Segmentation}
%
%
%

\author{Xiaolin~Zhang,
        Yunchao~Wei,
        Yi~Yang,
        and~Thomas~S.~Huang,~\IEEEmembership{Life~Fellow,~IEEE}
\thanks{X. Zhang, Y. Wei, Y. Yang are with the ReLER lab, Centre for Artificial Intelligence, University of Technology Sydney, Ultimo, NSW 2007, Australia.
(~e-mail: Xiaolin.Zhang-3@student.uts.edu.au, Yunchao.Wei@uts.edu.au, Yi.Yang@uts.edu.au~)}
\thanks{
T.~S.~Huang is with the Department of Electrical and Computer Engineering and Beckman Institute, University of Illinois at Urbana-Champaign, Urbana, IL, 61901 USA. (~e-mail:t-huang1@illinois.edu~)}
}

\markboth{IEEE Transactions on Cybernetics}%
{Shell \MakeLowercase{\textit{et al.}}: Bare Demo of IEEEtran.cls for IEEE Journals}

\maketitle

\begin{abstract}
One-shot image semantic segmentation poses a challenging task of recognizing the object regions from unseen categories with only one annotated example as supervision.
In this paper, we propose a simple yet effective Similarity Guidance network to tackle the One-shot (SG-One) segmentation problem. 
We aim at predicting the segmentation mask of a query image with the reference to one densely labeled support image of the same category.
To obtain the robust representative feature of the support image, we firstly adopt a masked average pooling strategy for producing the guidance features by only taking the pixels belonging to the support image into account.
We then leverage the cosine similarity to build the relationship between the guidance features and features of pixels from the query image. 
In this way, the possibilities embedded in the produced similarity maps can be adapted to guide the process of segmenting objects. 
Furthermore, our SG-One is a unified framework which can efficiently process both support and query images within one network and be learned in an end-to-end manner.
We conduct extensive experiments on Pascal VOC 2012. In particular, our SG-One achieves the mIoU score of 46.3\%, surpassing the baseline methods.

\end{abstract}

\begin{IEEEkeywords}
Few-shot Learning, Image Segmentation, Neural Networks, Siamese Network
\end{IEEEkeywords}

%
\IEEEpeerreviewmaketitle

\section{Introduction}
\IEEEPARstart{O}{bject}
Semantic Segmentation (OSS) aims at predicting the class label of each pixel.
Deep neural networks have achieved tremendous success on the OSS tasks, such as U-net~\cite{ronneberger2015u}, FCN~\cite{2015-long} and Mask R-CNN~\cite{he2017mask}.
However, these algorithms trained with full annotations require many investments to the expensive labeling tasks.
To reduce the budget, a promising alternative approach is to apply weak annotations for learning a decent network of segmentation.
For example, previous works have implemented image-level labels~\cite{wei2018revisiting,wei2017object,wei2015stc}, scribbles~\cite{lin2016scribblesup,tang2018normalized,wang2019boundary}, bounding boxes~\cite{khoreva2017simple,dai2015boxsup} and points~\cite{bearman2016s,tang2018weakly,qian2019weakly} as cheaper supervision information.
Whereas the main disadvantage of these weakly supervised methods is the lack of the ability for generalizing the learned models to unseen classes.
For instance, if a network is trained to segment dogs using thousands of images containing various breeds of dogs, it will not be able to segment bikes without retraining the network using many images containing bikes.

In contrast, humans are very good at recognizing things with a few guidance.
For instance, it is very easy for a child to recognize various breeds of dogs with the reference to only one picture of a dog.
Inspired by this, one-shot learning dedicates to imitating this powerful ability of human beings.
In other words, one-shot learning targets to recognize new objects according to only one annotated example.
This is a great challenge for the standard learning methodology.
Instead of using tremendous annotated instances to learn the characteristic patterns of a specific category, our target is to learn one-shot networks to generalize to unseen classes with only one densely annotated example.

\begin{figure}[t]
  \centering
  \includegraphics[width=.5\textwidth]{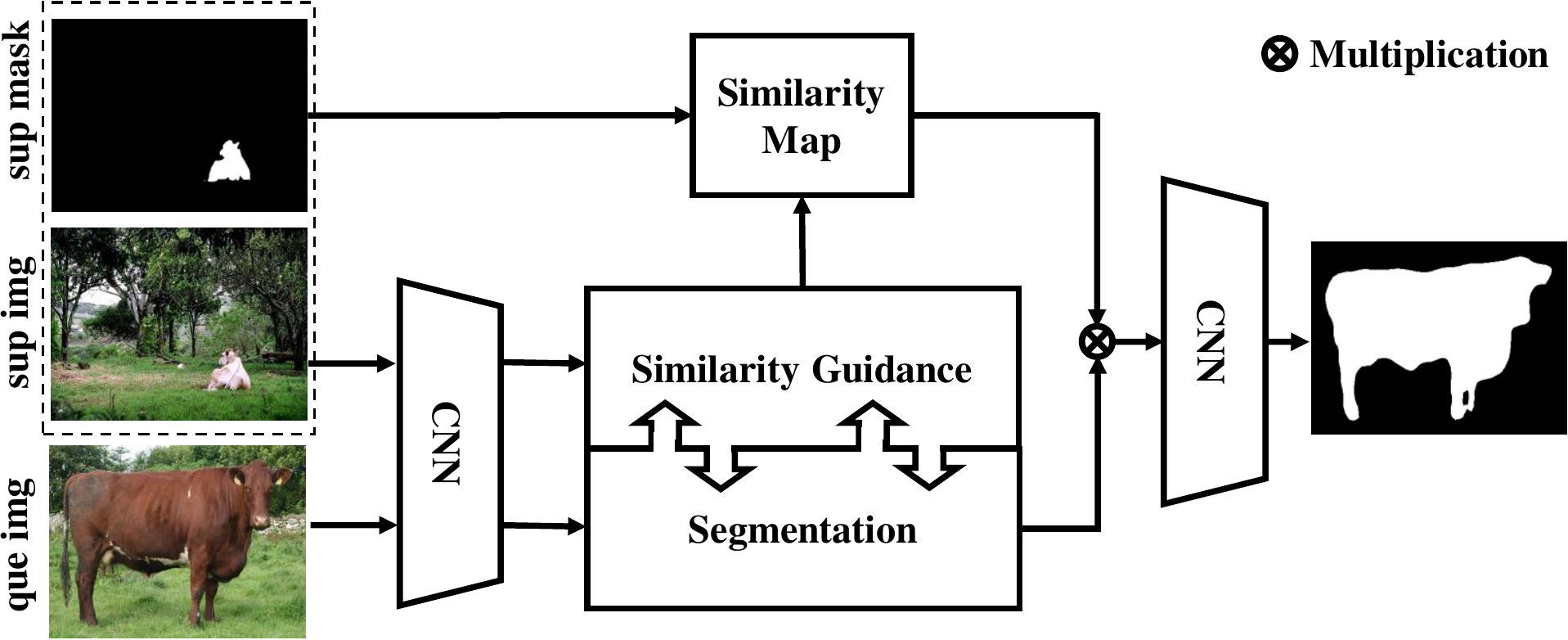}
  \caption{An overview of the proposed SG-One approach for testing a new class.
  Given a query image of an unseen category, \eg. cow, its semantic object is precisely segmented with the reference to only one annotated example of this category.
  \vspace{-10pt}
  }\label{fig-0}
\end{figure}
Concretely, one-shot image segmentation is to discover the object pixels of a query image with the reference to only one support image.
The target objects in the support image is densely annotated.
Current existing methods~\cite{shaban2017one,rakelly2018conditional} are all based on the Siamese framework~\cite{koch2015siamese}.
Briefly, a pair of parallel networks is trained for extracting the features of labeled support images and query images, separately.
These features are then fused to generate the probability maps of the target objects.
The purpose of the network is actually to learn the relationship between the annotated support image and the query image within the highlevel feature space.
These methods provide an advantage that the trained parameters of observed classes can be directly utilized for testing unseen classes without finetuning.
Nevertheless, there are some weaknesses with these methods:
1) The parameters of using the two parallel networks are redundant, which is prone to overfitting and leading to the waste of computational resources;
2) combining the features of support and query images by mere multiplication is inadequate for guiding the query network to learn high-quality segmentation masks.

To overcome the above mentioned weaknesses, we propose a \textbf{S}imilarity \textbf{G}uidance Network for \textbf{One}-Shot Semantic Segmentation (\textbf{SG-One}) in this paper.
The fundamental idea of SG-One is to guide the segmentation process by effectively incorporating the pixel-wise similarities between the features of support objects and query images.
Particularly, we firstly extract the highlevel feature maps of the input support and query images. Highlevel features are usually abstract and the embeddings of pixels belonging to the same category objects are close.
The embeddings with respect to the background pixels are usually depressed and these embeddings are distant from the object embeddings.
Therefore, we propose to obtain the representative vectors from support images by applying the masked average pooling operation. 
The masked average pooling operation can extract the object-related features by excluding the influences from background noises.
Then, we get the guidance maps by calculating cosine similarities between the representative vectors and the features of query images at each pixel.
The feature vectors corresponding to the pixels of objects in query images are close to the representative vectors extracted from support images, hence the scores in the guidance maps will be high.
Otherwise, the scores in guidance maps will be low if the pixels belonging to the background.
The generated guidance maps are applied to supply the guidance information of desired regions to the segmentation process. 
In detail, the position-wise feature vectors of query images are multiplied by the corresponding similarity values.
Such a strategy can effectively contribute to activating the target object regions of query images following the guidance of support images and their masks.  
Furthermore, we adopt a unified network for producing similarity guidance and predicting segmentation masks of query images.
Such a network is more capable of generalizing to unseen classes than the previous methods~\cite{shaban2017one, rakelly2018conditional}.

Our approach offers multiple appealing advantages over the previous state-of-the-arts, \eg OSLSM~\cite{shaban2017one} and co-FCN~\cite{rakelly2018conditional}.
First, OSLSM and co-FCN incorporate the segmentation masks of support images by changing the input structure of the network or the statistic distribution of input images.
Differently, we extract the representative vector from the intermediate feature maps with the masked average pooling operation instead of changing the inputs.
Our approach does not harm the input structure of the network, nor harm the statistics of input data. 
Averaging only the object regions can avoid the influences from the background.
Otherwise, when the background pixels dominate, the learned features will bias towards the background contents.
Second, OSLSM and co-FCN directly multiply the representative vector to feature maps of query images for predicting the segmentation masks.
SG-One calculates the similarities between the representative vector and the features at each pixel of query images, and the similarity maps are employed to guide the segmentation branch for finding the target object regions.
Our method is superior in the process of segmenting the query images. 
Third, both OSLSM and co-FCN adopt a pair of VGGnet-16 networks for processing support and query images, separately.
We employ a unified network to process them simultaneously.
The unified network utilizes much fewer parameters, so as to reduce the computational burden and increase the ability to generalize to new classes in testing.

The overview of SG-One is illustrated in Figure~\ref{fig-0}. 
We apply two branches, \ie, similarity guidance branch, and segmentation branch, to produce the guidance maps and segmentation masks.
We forward both the support and query images through the guidance branch to calculate the similarity maps.
The features of query images also forward through the segmentation branch for prediction segmentation masks.
The similarity maps act as guidance attention maps where the target regions have higher scores while the background regions have lower scores.
The segmentation process is guided by the similarity maps for getting the precise target regions.
After the training phase, the SG-One network can predict the segmentation masks of a new class without changing the parameters. 
For example, a query image of an unseen class, \eg, cow, is processed to discover the pixels belong to the cow with only one annotated support image provided.

To sum up, our main contributions are three-fold:
\begin{itemize}
\item We propose to produce robust object-related representative vectors using masked average pooling for incorporating contextual information without changing the input structure of networks.  
\item We produce the pixel-wise guidance using cosine similarities between representative vectors and query features for predicting the segmentation masks. 
\item We propose a unified network for processing support and query images. Our network achieves the cross-validate mIoU of 46.3\% on the PASCAL-5\textsuperscript{i} dataset in one-shot segmentation setting, surpassing the baseline methods.
\end{itemize}

\begin{figure*}[htp]
  \centering
  \includegraphics[width=1.0\textwidth]{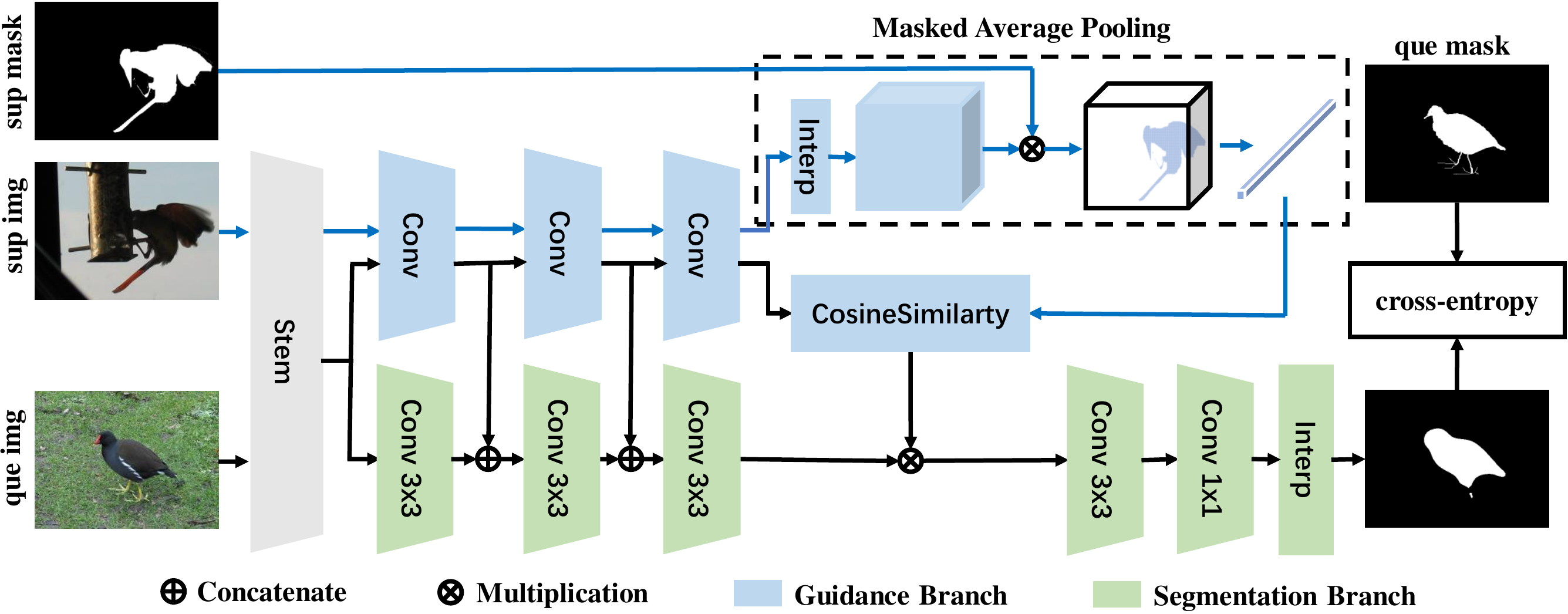}
  \caption{The network of the proposed SG-One approach. A query image and a labeled support image are fed into the network. Guidance Branch is to extract the representative vector of the target object in the support image. Segmentation Branch is to predict the segmentation masks of the query image. We calculate the cosine distance between the vector and the intermediate features of the query image. The CosineSimilarty maps are then employed to guide the segmentation process. 
  The \textit{blue arrows} indicate data streams of support images, while the \textit{black} are for query images.
  Stem is the \textit{conv1} to \textit{conv3} of VGG16. Interp refers to the bilinear interpolation operation. Conv is a convoluational block. Conv $k\times k$ is the convolutional filter with a kernel size of $k\times k$}.
  \label{fig-1}
\vspace{-3mm}
\end{figure*}
\section{Related Work}
\textbf{Object semantic segmentation} (OSS) aims at classifying every pixel in a given image for distinguishing different objects or contents. 
OSS with dense annotations as supervision has achieved great success in precisely identifying various kinds of objects~\cite{husain2014consistent,yang2014scale, peng2015high,nie20183,wang2018hybrid,ben2019coupled,cheng2019spgnet,huang2018ccnet}.
Recently, most of the works with impressive performance are based on deep convolutional networks. 
FCN~\cite{2015-long} and U-Net~\cite{ronneberger2015u} abandon fully connected layers and propose to only use convolutional layers for preserving relative positions of pixels.
Based on the advantages of FCN, DeepLab proposed by Chen \etal~\cite{chen2014semantic,chen2018encoder}, is one of the best algorithms for segmentation.
It employs dilated convolution operations to increase the receptive field, and meanwhile to save parameters in comparison with the large convolutional kernel methods.
He~\etal~\cite{he2017mask} proposes segmentation masks and detection bounding boxes can be predicted simultaneously using a unified network.

\textbf{Weakly object segmentation} seeks an alternative approach for segmentation to reduce the expenses in labeling segmentation masks~\cite{wang2018weakly,wang2017instance,liu2019weakly,qian2019weakly,hou2018self,jiang2019integral}.
Zhou~\cite{zhou2015cnnlocalization} and Zhang~\cite{zhang2018adversarial,zhang2018self} propose to discover precise object regions using a classification network with only image-level labels.
Wei~\cite{wei2018revisiting,wei2017object} apply a two-stage mechanism to predict segmentation masks.
Concretely, confident regions of objects and background are firstly extracted via the methods of Zhou~\etal or Zhang~\etal
Then, a segmentation network, such as DeepLab, can be trained for segmenting the target regions. 
An alternative weakly segmentation approach is to use scribble lines to indicate the rough positions of objects and background.
Lin~\etal~\cite{lin2016scribblesup} and Tang~\etal~\cite{tang2018normalized} adopted spectral clustering methods to distinguish the object pixels according to the similarity of adjacent pixels and ground-truth scribble lines. 

\textbf{Video Object Segmentation} is also a challenging problem of segmenting a specific object in a video clip given merely the annotation of the first frame~\cite{davis2017}. The categories of training and testing sets are disjoint, which makes the task similar to our one-shot image segmentation.
OSVOS~\cite{caelles2017one} adopts a direct approach of learning a segmentation network on the training set, and then finetunes the trained network on the augmented first frames of testing sets. 
Although OSVOS achieves good segmentation performance, the main drawback is the latency of finetuning during testing a new video clip. 
PLM~\cite{shin2017pixel} applies a more sophisticated network to learn better feature embeddings by involving intermediate feature maps of both search and query frames.
A simple crop method is also applied by estimating the approximate location of target objects according to the relationship between successive frames. 
SegFlow~\cite{cheng2017segflow} leverages optical flow of moving objects to assist the segmentation process.
Flownet~\cite{dosovitskiy2015flownet} are internally embedded to the framework of SegFlow, and updated in an end-to-end way.
Consequently, the segmentation network and flow network can benefit from each other to learn better segmentation masks as well as optical flows.
VideoMatch~\cite{hu2018videomatch} learns the represents of both foreground and background regions, because the successive video clips usually maintain similar or static background environments. 
Therefore, the learned robust represents can be easily applied to retrieve the target object regions of query images.

\textbf{Few-shot learning} algorithms dedicates to distinguishing the patterns of classes or objects with only a few labeled samples~\cite{wu2019progressive,he2019asymptotic}. 
Networks should generalize to recognize new objects with few images based on the parameters of base models.
The base models are trained using entirely different classes without any overlaps with the testing classes.
Finn~\etal~\cite{finn2017model} tries to learn some internal transferable representations, and these representations are broadly applicable to various tasks.
Vinyals~\etal~\cite{vinyals2016matching} and Annadani~\etal~\cite{annadani2018preserving} propose to learn  embedding vectors. 
The vectors of the same categories are close, while the vectors of the different categories are apart.

\section{Methodology}
\subsection{Problem Definition}
Suppose we have three datasets: a training set $L_{train}=\{(I_i, Y_i)\}_{i=1}^{N_{train}}$, a support set $L_{support}=\{(I_i, Y_i)\}_{i=1}^{N_{support}}$ and a testing set $L_{test}=\{I_i\}_{i=1}^{N_{test}}$, where $I_i$ is an image, $Y_i$ is the corresponding segmentation mask and $N$ is the number of images in each set.
Both the support set and training set have annotated segmentation masks.
The support set and testing set share the same types of objects which are disjoint with the training set.
We denote $l\in Y$ as the semantic class of the mask $Y$. Therefore, we have $\{l_{train}\} \cap \{l_{support}\}=\emptyset$, 
where $\cap$ denotes the intersection of the two sets.
If there are $K$ annotated images for each class, the target few-shot problem is named $K$-shot.
Our purpose is to train a network on the training set $L_{train}$, which can precisely predict segmentation masks $\hat{Y}_{test}$ on testing set $L_{test}$ according to the reference of the support set $L_{support}$.
Specifically, the predicted masks contains two classes, \ie, object class and background. 
If the objects in query images share the same category label with the annotated objects in support images, the corresponding values in the predicted masks are supposed to be $1$ for indicating object pixels. Otherwise, the corresponding values should be $0$ for indicating background pixels.

In order to better learn the connection between the support and testing set, we mimic this mechanism in the training process.
For a query image $I_i$, we construct a pair $\{(I_i, Y_i), (I_j, Y_j)\}$ by randomly selecting a support image $I_j$ whose mask $Y_j$ has the same semantic class as $Y_i$.
We are supposed to estimate the segmentation mask $\hat{Y}_i$ with a function $\hat{Y}_j = f_{\theta}((I_i,Y_i), I_j)$, where $\theta$ is the parameters of the function.
In testing, $(I_i, Y_i)$ is picked from the support set $L_{support}$ and $I_j$ is an image from testing set $L_{test}$. 

\subsection{Proposed Model}\label{model}
In this section, we firstly present the masked average pooling operation for extracting the object-related representative vector of annotated support images.
Then, the similarity guidance method is introduced for combining the representative vectors and features of query images. 
The generated similarity guidance maps supply the information for precisely predicting the segmentation masks.

\noindent \textbf{Masked Average Pooling}
The pairs of support images and their masks are usually encoded into representative vectors.
OSLSM~\cite{shaban2017one} proposes to erase the background pixels from the support images by multiplying the binary masks to support images.
co-FCN~\cite{rakelly2018conditional} proposes to construct the input block of five channels by concatenating the support images with their positive and negative masks.
However, there are two disadvantages of the two methods. 
First, erasing the background pixels to zeros will change the statistic distribution of the support image set.
If we apply a unified network to process both the query images and the erased support images, the variance of the input data will greatly increase.
Second, concatenating the support images with their masks~\cite{rakelly2018conditional} breaks the input structure of the network, which will also prevent the implementation of a unified network.

We propose to employ masked average pooling for extracting the representative vectors of support objects.
Suppose we have a support RGB image $I\in R^{3\times w \times h}$ and its binary segmentation mask $Y \in \{0,1\}^{w\times h}$, where $w$ and $h$ are the width and height of the image.
If the output feature maps of $I$ is $F'\in R^{c\times w' \times h'}$, where $c$ is the number of channels, $w'$ and $h'$ are width and height of the feature maps.
We firstly resize the feature maps to the same size as the mask $Y$ via bilinear interpolation.
We denote the resized feature maps as $F\in R^{c\times w \times h}$.
Then, the $i_{th}$ element $v_{i}$ of the representative vector $\mathbf{v}$ is computed by averaging the pixels within the object regions on the $i_{th}$ feature map.
\begin{equation}\label{eq1}
v_i = \frac{\sum_{x=1,y=1}^{w,h} Y_{x,y}*F_{i,x,y}}{\sum_{x=1,y=1}^{w,h} Y_{x,y}},
\end{equation}
As the discussion in FCN~\cite{2015-long}, fully convolutional networks are able to preserve the relative positions of input pixels.
Therefore, through masked average pooling, we expect to extract the features of object regions while disregarding the background contents.
Also, we argue that the input of contextual regions in our method is helpful to learn better object features.
This statement has been discussed in DeepLab~\cite{chen2014semantic} which incorporates contextual information using dilated convolutions.
Masked average pooling keeps the input structure of the network unchanged, which enables us to process both the support and query images within a unified network.

\noindent \textbf{Similarity Guidance}
One-shot semantic segmentation aims to segment the target object within query images given a support image of the reference object.
As we have discussed, the masked average pooling method is employed to extract the representative vector $\mathbf{v}=(v_1,v_2,...,v_c)$ of the reference object, where $c$ is the number of channels.
Suppose the feature maps of a query image $I^{que}$ is $F^{que}\in R^{c\times w' \times h'}$.
We employ the cosine distance to measure the similarity between the representative vector $\mathbf{v}$ and each pixel within $F^{que}$ following Eq.~\eqref{eq2}
\begin{equation}\label{eq2}
s_{x,y} = \frac{\mathbf{v}* F^{que}_{x,y}}{||\mathbf{v}||_2 * ||F^{que}_{x,y}||_2},
\end{equation}
where $s_{x,y} \in [-1,1]$ is the similarity value at the pixel $(x,y)$.
$F^{que}_{x,y} \in R^{c\times 1}$ is the feature vector of query image at the pixel $(x,y)$.
As a result, the similarity map $S$ integrates the features of the support object and the query image.
We use the map $S=\{s_{x,y}\}$ as guidance to teach the segmentation branch to discover the desired object regions.
We do not explicitly optimize the cosine similarity. 
In particular, we element-wisely multiply the similarity guidance map to the feature maps of query images from segmentation branch.
Then, we optimize the guided feature maps to fit the corresponding ground truth masks.

\subsection{Similarity Guidance Method}

Figure~\ref{fig-1} depicts the structure of our proposed model.
SG-One includes three components,~\ie, Stem, Similarity Guidance Branch, and Segmentation Branch.
Different components have different structures and functionalities.
Stem is a fully convolutional network for extracting intermediate features of both support and query images. 

Similarity Guidance Branch is fed the extracted features of both query and support images.
We apply this branch to produce the similarity guidance maps by combining the features of reference objects with the features of query images.
For the features of support images, we implement three convolutional blocks to extract the highly abstract and semantic features, followed by a masked averaged pooling layer to obtain representative vectors.
The extracted representative vectors of support images are expected to contain the high-level semantic features of a specific object. 
For the features of query images, we reuse the three blocks and employ the cosine similarity layer to calculate the closeness between the representative vector and the features at each pixel of the query images.  

Segmentation branch is for discovering the target object regions of query images with the guidance of the generated similarity maps.
We employ three convolutional layers with the kernel size of $3 \times 3$ to obtain the features for segmentation.
The inputs of the last two convolutional layers are concatenated with the paralleling feature maps from Similarity Guidance Branch.
Through the concatenation, Segmentation Branch can borrow features from the paralleling branch, and these two branches can communicate information during the forward and backward stages.
We fuse the generated features with the similarity guidance maps by multiplication at each pixel. 
Finally, the fused features are processed by two convolutional layers with the kernel size of $3 \times 3$ and $1 \times 1$, followed by a bilinear interpolation layer.
The network finally classifies each pixel to be the same class with support images or to be background.
We employ the cross-entropy loss function to optimize the network in an end-to-end manner.

\noindent \textbf{One-Shot Testing}
One annotated support image for each unseen category is provided as guidance to segment the target semantic objects of query images.
We do not need to finetune or change any parameters of the entire network.
We only need to forward the query and support images through the network for generating the expected segmentation masks.

\noindent \textbf{K-Shot Testing}
Suppose there are $K (K>1)$  support images $I^i_{sup}, i=\{1,2...,K\}$ for each new category, we propose to segment the query image $I_{que}$ using two approaches.
The first one is to ensemble the segmentation masks corresponding to the $K$ support images following OSLSM~\cite{shaban2017one} and co-FCN~\cite{rakelly2018conditional} based on Eq.~\eqref{eq3}
\begin{equation}\label{eq3}
\hat{Y}_{x,y} =  max(\hat{Y}_{x,y}^{1}, \hat{Y}_{x,y}^{2},...,\hat{Y}_{x,y}^{K}),
\end{equation}
where $\hat{Y}^{i}_{x,y}, i=\{1,2,...,K\}$ is the predicted semantic label of the pixel at $(x,y)$ corresponding to the support image $I^i_{sup}$.
Another approach is to average the $K$ representative vectors, and then use the averaged vector to guide the segmentation process.
It is notable that we do not need to retrain the network using $K$-shot support images.
We use the trained network in a one-shot manner to test the segmentation performance using $K$-shot support images.

\section{Experiments}
\subsection{Dataset and Metric}
Following the evaluation method of the previous methods OSLSM~\cite{shaban2017one} and co-FCN~\cite{rakelly2018conditional}, we create the \textbf{PASCAL-5\textsuperscript{i}} using the PASCAL VOC 2012 dataset~\cite{2010-pascal} and the extended SDS dataset~\cite{hariharan2011semantic}. 
For the 20 object categories in PASCAL VOC, we use cross-validation method to evaluate the proposed model by sampling five classes as test categories $L_{test}=\{4i+1,...,4i+5\}$ in Table~\ref{tab1}, where $i$ is the fold number, while the left 15 classes are the training label-set $L_{train}$.
We follow the same procedure of the baseline methods~\eg OSLSM~\cite{shaban2017one} to build the training and testing set. 
Particularly, we randomly sample image pairs from the training set.
Each image pair have one common category label.
One image is fed into the network as a support image accompanied by its annotated mask.
Another image is treated as a query image and its mask is applied to calculate the loss.
For a fair comparison, we use the same test set as OSLSM~\cite{shaban2017one}, which has 1,000 support-query tuples for each fold.

\begin{table}[t]\setlength{\tabcolsep}{19.5pt}
  \centering
  \caption{Testing classes for 4-fold cross-validation test. The training classes of PASCAL-5\textsuperscript{i}, i=\{0,1,2,3\} are disjoint with the testing classes.}\label{tab1}
  \begin{tabular}{l|c}
    \hline
    \hline
    \textbf{Dataset} & \textbf{Test classes} \\
    \hline
    PASCAL-5\textsuperscript{0} & aeroplane,bicycle,bird,boat,bottle \\
    PASCAL-5\textsuperscript{1} & bus,car,cat,chair,cow \\
    PASCAL-5\textsuperscript{2} & diningtable,dog,horse,motorbike,person \\
    PASCAL-5\textsuperscript{3} & potted plant,sheep,sofa,train,tv/monitor \\
    \hline
    \hline
  \end{tabular}
  \vspace{-10pt}
\end{table}

Suppose the predicted segmentation mask is $\{\hat{M}\}_{i=1}^{N_{test}}$ and the corresponding ground-truth annotation is $\{M\}_{i=1}^{N_{test}}$, given a specific class $l$.
We define the Intersection over Union ($IoU_l$) of class $l$ as $\frac{TP_l}{TP_l+FP_l+FN_l}$, where the $TP, FP$ and $FN$ are the number of true positives, false positives and false negatives of the predicted masks.
The mIoU is the average of IoUs over different classes, \ie $(1/n_l)\sum_l IoU_l$, where $n_l$ is the number of testing classes.
We report the averaged mIoU on the four cross-validation datasets.

\subsection{Implementation details}
We implement the proposed approach based on the VGG-16 network following the previous works~\cite{shaban2017one,rakelly2018conditional}.
Stem takes the input of RGB images to extract middle-level features, and downsamples the images by the scale of 8.
We use the first three blocks of the VGG-16 network as Stem.
For the first two convolutional blocks of Similarity Guidance Branch, we adopt the structure of \textit{conv4} and \textit{conv5} of VGGnet-16 and remove the \textit{maxpooling} layers to maintain the resolution of feature maps.
One \textit{conv3$\times$3} layer of 512 channels are added on the top without using $ReLU$ after this layer. 
The following module is masked average pooling to extract the representative vector of support images.
In Segmentation Branch, all of the convolutional layers with $3 \times 3$ kernel size have 128 channels.
The last layer of \textit{conv1$\times$1} kernel has two channels corresponding to categories of either object or background. 
All of the convolutional layers except for the third and the last one are followed by a $ReLU$ layer.
We will justify this choice in section~\ref{ablation}.

Following the baseline methods~\cite{shaban2017one,rakelly2018conditional}, we use the pre-trained weights on ILSVRC~\cite{ILSVRC15}. 
All input images remain the original size without any data augmentations. 
The support and query images are fed into the network, simultaneously.
The difference from them is the support image just go through the guidance branch to obtain the representative vector.
The query images goes through both the guidance branch for calculating the guidance maps and go through the segmentation branch for predicting the segmentation masks.
We implement the network using PyTorch~\cite{pytorch}.
We train the network with the learning rate of $1e-5$.
The batch size is $1$, and the weight decay is 0.0005.
We adopt the SGD optimizer with the momentum of 0.9.
All networks are trained and tested on NVIDIA TITAN X GPUs with 12 GB memory.
Our source code is available at \url{https://github.com/xiaomengyc/SG-One}.

\begin{table*}[ht]\setlength{\tabcolsep}{18pt}
  \centering
  \caption{Mean IoU results of one-shot segmentation on the PASCAL-5\textsuperscript{i} dataset. The best  results are in bold.}\label{comp_one}
  \begin{tabular}{l|cccc|c}
    \hline
    \hline
    \textbf{Methods\footnotemark} 
    & \textbf{PASCAL-5\textsuperscript{0}} & \textbf{PASCAL-5\textsuperscript{1}} & \textbf{PASCAL-5\textsuperscript{2}} & \textbf{PASCAL-5\textsuperscript{3}} & \textbf{Mean} \\
    \hline
    1-NN &  25.3 & 44.9 & \textbf{41.7} & 18.4  & 32.6\\
    LogReg & 26.9 & 42.9 & 37.1 & 18.4 & 31.4\\
    Siamese & 28.1 & 39.9 & 31.8 & 25.8 & 31.4\\
    OSVOS~\cite{caelles2017one} & 24.9 & 38.8 & 36.5 & 30.1 & 32.6\\
    OSLSM~\cite{shaban2017one} &  33.6 & 55.3 & 40.9 & 33.5 & 40.8\\
    co-FCN~\cite{rakelly2018conditional} &  36.7  & 50.6 & 44.9 & 32.4 & 41.1 \\
    SG-One(Ours) &  \textbf{40.2} & \textbf{58.4}  & \textbf{48.4}  & \textbf{38.4} & \textbf{46.3} \\
    \hline
    \hline
  \end{tabular}
  \vspace{-10pt}
\end{table*}

\begin{table*}[ht]\setlength{\tabcolsep}{17pt}
  \centering
  \caption{Mean IoU results of five-shot segmentation on the PASCAL-5\textsuperscript{i} dataset. The best results are in bold.}\label{comp_five}
  \begin{tabular}{l|cccc|c}
    \hline
    \hline
    \textbf{Methods} & \textbf{PASCAL-5\textsuperscript{0}} & \textbf{PASCAL-5\textsuperscript{1}} & \textbf{PASCAL-5\textsuperscript{2}} & \textbf{PASCAL-5\textsuperscript{3}} & \textbf{Mean} \\
    \hline
    1-NN & 34.5 & 53.0 & 46.9 & 25.6 & 40.0 \\
    LogReg & 35.9 & 51.6 & 44.5 & 25.6 & 39.3\\
    OSLSM~\cite{shaban2017one} &  35.9 & 58.1 & 42.7 & 39.1 & 43.9\\
    co-FCN~\cite{rakelly2018conditional} &  37.5 & 	50.0 & 44.1 & 33.9 & 41.4\\
    SG-One-max(Ours) & 40.8  & 57.2 & 46.0 & 38.5  & 46.2\\
    SG-One-avg(Ours) & \textbf{41.9}  & \textbf{58.6} &  \textbf{48.6} & \textbf{39.4} & \textbf{47.1} \\
    \hline
    \hline
  \end{tabular}
  \vspace{-10pt}
\end{table*}

\begin{table}[ht]\setlength{\tabcolsep}{22pt}
  \centering
  \caption{Mean IoU results of one-shot segmentation regarding the evaluation method in co-FCN~\cite{rakelly2018conditional} and PL+SEG~\cite{dong2018few}}\label{metric2-tab}
  \begin{tabular}{l|cc}
    \hline
    \hline
    \textbf{Methods} & \textbf{one-shot} & \textbf{five-shot} \\
    \hline
    FG-BG~\cite{rakelly2018conditional} &  55.1 & 55.6 \\
    OSLSM~\cite{shaban2017one} &  55.2 & - \\
    co-FCN~\cite{rakelly2018conditional} &  60.1 & 	60.8 \\
    PL+SEG~\cite{dong2018few} &  61.2  & 62.3 \\
    SG-One(Ours) & \textbf{63.1}  & \textbf{65.9}\\
    \hline
    \hline
  \end{tabular}
  \vspace{-10pt}
\end{table}

\begin{figure*}
  \centering
  \includegraphics[width=1.0\textwidth]{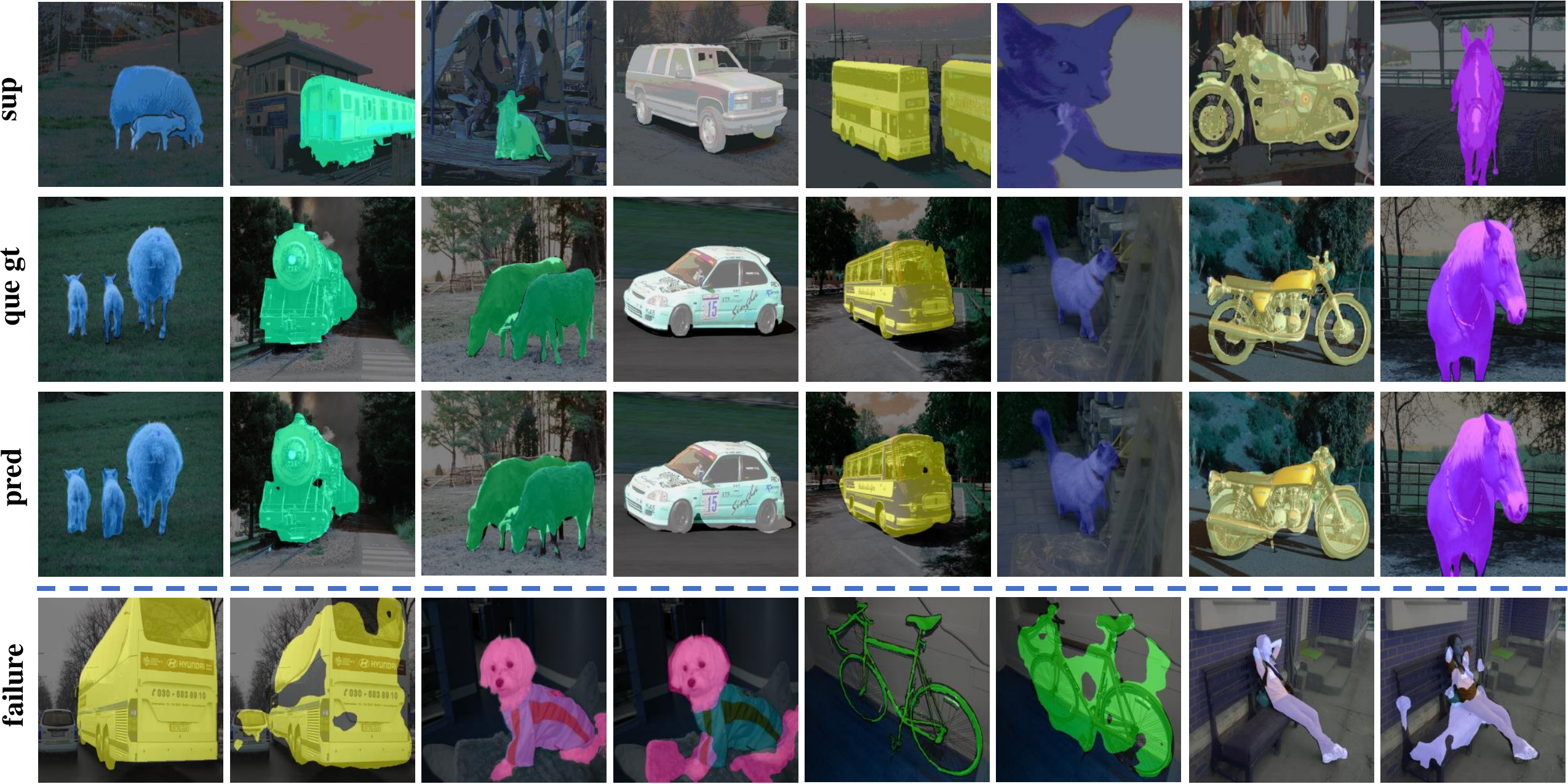}
  \caption{Segmentation results on unseen classes with the guidance of support images. For the failure pairs, the ground-truth is on the \textit{left} side while the predicted is on the \textit{right} side.}\label{show-one}
  \vspace{-12pt}
\end{figure*}

\begin{figure}
  \centering
  \includegraphics[width=0.45\textwidth]{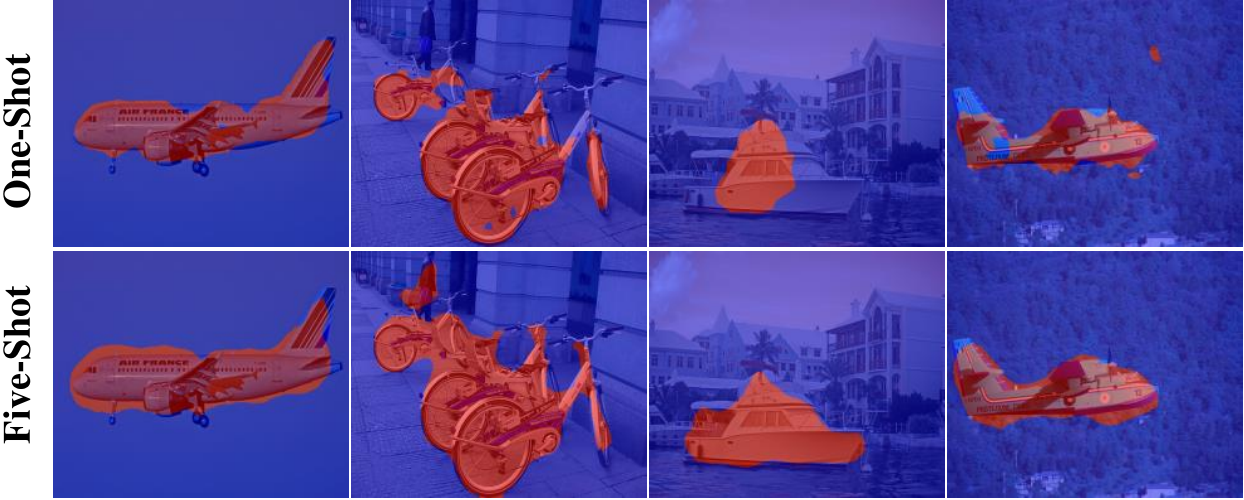}
  \caption{Qualititive illustration of the one-shot and five-shot segmentation.}\label{show-1to5}
  \vspace{-12pt}
\end{figure}

\begin{figure*}
  \centering
  \includegraphics[width=1.0\textwidth]{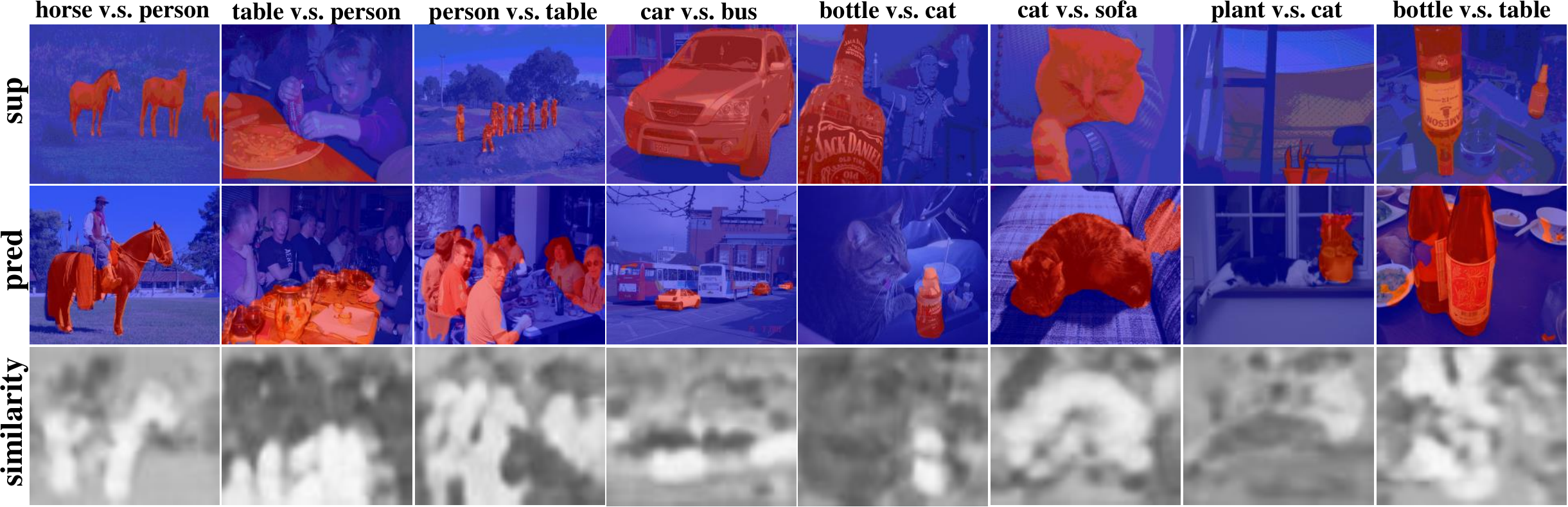}
  \caption{Similarity maps of different categories. With the reference to the support objects, the objects in the query images of the same categories will be highlighted, while the distracting objects and the background are depressed. The predicted mask can precisely segment the target objects under the guidance of the similarity maps.}\label{simi_map}
  \vspace{-12pt}
\end{figure*}

\subsection{Comparison}
\noindent \textbf{One-shot}
Table~\ref{comp_one} compares the proposed SG-One approach with baseline methods in one-shot semantic segmentation. 
We observe that our method outperforms all baseline models.
The mIoU of our approach on the four divisions achieves 46.3\%, which is significantly better than co-FCN by 5.2\% and OSLSM by 5.5\%.
Compared to the baselines, SG-One earns the largest gain of 7.8\% on PASCAL-5\textsuperscript{1}, where the testing classes are \textit{bus, car, cat, chair} and \textit{cow}.
co-FCN~\cite{rakelly2018conditional} constructs the input of the support network by concatenating the support images, positive and negative masks, and it obtains 41.1\%.
OSLSM~\cite{shaban2017one} proposed to feed only the object pixels as input by masking out the background regions, and this method obtains 40.8\%. 

OSVOS~\cite{caelles2017one} adopts a strategy of finetuning the network using the support samples in testing, and it achieves only 32.6\%.
To summarize, SG-One can effectively predict segmentation masks on new classes without changing the parameters.
Our similarity guidance method is better than the baseline methods in incorporating the support objects for segmenting unseen objects.

Figure~\ref{show-one} shows the one-shot segmentation results using SG-One on unseen classes.
We observe that SG-One can precisely distinguish the object regions from the background with the guidance of the support images, even if some support images and query images do not share much appearance similarities.
We also show some failure cases to benefit the future researches.
We ascribe the failure to 1) the target object regions are too similar to background pixels, \eg the side of the bus and the car; 
2) the target region have very uncommon features with the discovered discriminative regions, \eg the vest of the dog, which may far distant with the representative feature of support objects.

\footnotetext{The details of baseline methods~\eg 1-NN, LogReg and Siamese, refer to OSLSM~\cite{shaban2017one}.
The results for co-FCN~\cite{rakelly2018conditional} are for our re-implemented version.
Table~\ref{metric2-tab} reports the evaluation results regarding the same metric adopted in~\cite{rakelly2018conditional} for a fairer comparison.
}

Figure~\ref{simi_map} illustrates the similarity maps of cosine distance between the support objects and the query images.
We try to segment the objects of the query images in the second row corresponding to the annotated objects of the support images in the first row.
Note there exist distracting classes in the given query images. 
We only expect to segment the objects whose categories are consistent with the support images.
With the reference to the extracted features of support objects, the corresponding regions in the query images are highlighted while the distracting regions and the background are depressed.
The predicted masks can be precisely predicted with the guidance to the similairty maps.

\noindent \textbf{Five-shot}
Table~\ref{comp_five} illustrates the five-shot segmentation results on the four divisions. 
As we have discussed, we apply two approaches to five-shot semantic segmentation.
The approach of averaging the representative vectors from the five support images achieves 47.1\% which significantly outperforms the current state-of-the-art, co-FCN, by 5.7\%.
This result is also better than the corresponding one-shot mIoU of 46.3\%.
Therefore, the averaged support vector has a better expressiveness of the features in guiding the segmentation process.
Another approach is to solely fuse the final segmentation results by combining all of the detected object pixels.
We do not observe any improvement of this approach, comparing to the one-shot result.
It is notable that we do not specifically train a new network for five-shot segmentation.
The trained network in a one-shot way is directly applied to predict the five-shot segmentation results.
Figure~\ref{show-1to5} compares the predicted segmentation masks of one-shot and five-shot.
The segmentation masks of five-shot are slightly better than that from one-shot prediction.
As we have also observed that five-shot testing can only improve the mIoU by $0.8$ which is a marginal growth. 
We think the reason for this phenomenon is that the highlevel features for different objects sharing the same class labels are very close.
Hence, averaging these features from different objects can only improve a little in terms of feature expressiveness which causes the five-shot increase is limited.
On the other side, our target of the similarity learning is exactly to produce the aligned features for each category.
So, the five-shot results can only improve a little under the current one-shot segmentation settings.

For a fairer comparison, we also evaluate the proposed model regarding the same metric in co-FCN~\cite{rakelly2018conditional} and PL+SEG~\cite{dong2018few}.
This metric firstly calculates the IoU of foreground and background, and then obtains the mean IoU of the foreground and background pixels.
We still report the averaged mIoU on the four cross-validation datasets.
Table~\ref{metric2-tab} compares SG-One with the baseline methods regarding this metric in terms of one-shot and five-shot semantic segmentation.
It is obvious that the proposed approach outperforms all previous baselines.
SG-One achieves 63.1\% of one-shot segmentation and 65.9\% of five-shot segmentation, while the most competitive baseline method PL+SEG only obtains 61.2\% and 62.3\%.
The proposed network is trained end-to-end, and our results do not require any pre-processing or post-processing steps. 

\vspace{-20pt}
\subsection{Multi-Class Segmentation}
We conduct experiments to verify the ability of SG-One in segmenting images with multiple classes.
We randomly select 1000 entries of the query and support images.
Query images may contain objects of multiple classes.
For each entry, we sample five annotated images from the five testing classes as support images.
For every query image, we predict its segmentation masks with the images of different support classes.
We fuse the segmentation masks of the five classes by comparing the classification scores.
The mIoU on the four datasets is 29.4\%.
We conduct the same experiment to test the co-FCN algorithm~\cite{rakelly2018conditional}, and co-FCN only obtains the mIoU of 11.5\%.
Therefore, we can tell that SG-One is much more robust in dealing with multi-cluass images, although it is trained with images of single class.

\subsection{Ablation Study}\label{ablation}
\noindent \textbf{Masked Average Pooling}
The masked average pooling method employed in the proposed SG-One network is superior in incorporating the guidance masks of support images. 
Shaban~\etal~\cite{shaban2017one} proposed to multiply the binary masks to the input support RGB images, so that the network can only extract features of target objects.
co-FCN~\cite{rakelly2018conditional} proposed by Rakelly~\etal concatenates the support RGB images with the corresponding positive masks, \ie object pixels are 1 while background pixels are 0, and negative binary masks \ie object pixels are 1 and background pixels are 0, constituting the inputs of 5 channels.
We follow the instructions of these two methods and compare with our masked average pooling approach.
Concretely, we firstly replace the masked average pooling layer by a global average pooling layer.
Then, we implement two networks.
1) \textbf{SG-One-masking} adopts the methods in OSLSM~\cite{shaban2017one}, in which support images are multiplied by the binary masks to solely keep the object regions. 
2) \textbf{SG-One-concatenate} adopts the methods in co-FCN~\cite{rakelly2018conditional}, in which we concatenate the positive and negative masks to the support images forming an input with 5 channels. 
We add an extra input block (VGGnet-16) with 5 convolutional channels for adapting concatenated inputs, while the rest networks are exactly the same as the compared networks.

Table~\ref{tab_input_process} compares the performance of different methods in processing support images and masks.
Our masked average pooling approach achieves the best results on every dataset.
The mIoU of the four datasets is 46.3\% using our method.
The masking method (SG-One-masking) proposed in OSLSM~\cite{shaban2017one} obtains 45.0\% of the mIoU.
The approach of co-FCN (SG-One-concat) only obtains 41.75\%, which ascribes the modification of the input structure of the network.
The modified input block cannot benefit from the pre-trained weights of processing low-level information.
We also implement a network using the general GAP layer to extract representative vectors instead of using the binary masks of the support images.
The network under such a setting achieves the mIoU of 42.2\%, which is inferior to the proposed MAP method.
So, it is necessary to mask out the pixels corresponding to the background regions for better representative vectors.
In total, we can conclude that 
1) a qualified method of using support masks is crucial for extracting high-quality object features; 
2) the proposed masked average pooling method provides a superior way to reuse the structure of well-designed classification network for extracting object features of support pairs;
3) networks with 5-channel input cannot benefit from the pre-trained weights and the extra input block cannot be jointly trained with the query images.
4) the masked average pooling layer has superior generalization ability in segmenting unseen classes.

\begin{table}\setlength{\tabcolsep}{8pt}
  \centering
  \caption{Comparison in different methods of extracting representative vectors. ($i$ denotes the PASCAL-5\textsuperscript{i} dataset.)}\label{tab_input_process}
  \begin{tabular}{l|cccc|c}
    \hline
    \hline
    \textbf{Methods} & $i=0$ & $i=1$ & $i=2$ & $i=3$ & Mean \\
    \hline
    SG-One-concat &  38.4 & 51.4 & 44.8 & 32.5 & 41.75 \\
    SG-One-masking &  41.9 & 55.3 & 47.4 & 35.5 & 45.0 \\
    SG-One-ours & \textbf{40.2} & \textbf{58.4}  & \textbf{48.4}  & \textbf{38.4}  & \textbf{46.3} \\
    \hline
    \hline
  \end{tabular}
  \vspace{-10pt}
\end{table}

\noindent \textbf{Guidance Similarity Generating Methods}
We adopt the cosine similarity to calculate the distance between the object feature vector and the feature maps of query images.
The definition of the cosine distance is to measure the angle between two vectors, and its range is in $[-1,1]$.
Correspondingly, we abandon the \textit{ReLU} layers after the third convolutional layers of both guidance and segmentation branches.
By doing so, we increase the variance of the cosine measurement, and the cosine similarity is not partly bounded in $[0,1]$, but in $[-1,1]$.
For comparison, we add the \textit{ReLU} layers after the third convolutional layers. 
The mIoU on the four datasets drops to 45.5\% comparing to the \textit{non-ReLU} approach of 46.3\%.

We also train a network using the \textit{2-norm} distance as the guidance, and obtain 30.7\% on the four datasets.
This result is far poor than the proposed cosine similarity method.
Hence, the \textit{2-norm} distance is not a good choice for guiding the query images to discover target object regions.

\noindent \textbf{The Unified Structure}
We adopt the proposed unified structure between the guidance and segmentation branches.
This structure can benefit each other during the forward and backward stages.
We implement two networks for illustrating the effectiveness of this structure.
First, we remove the first three convolutional layers of Segmentation Branch, and then multiply the guidance similarity maps directly to the feature maps from Similarity Guidance Branch.
The final mIoU of the four datasets decreases to 43.1\%.
Second, we cut off the connections between the two branches by removing the first and second concatenation operations between the two branches.
The final mIoU obtains 45.7\%.
Therefore, Segmentation Branch in our unified network is very necessary for getting high-quality segmentation masks.
Also, Segmentation Branch can borrow some information via the concatenation operation between the two branches. 

We also verify the functionality of the proposed unified network in the demand of computational resources and generalization ability.
In Table~\ref{tab_num_para}, we observe that our SG-One model has only 19.0M parameters, while it achieves the best segmentation results.
Following the methods in OSLSM~\cite{shaban2017one} and co-FCN~\cite{rakelly2018conditional}, we use a separate network (SG-One-separate) to process support images. 
This network has slightly more parameters (36.1M) than co-FCN(34.2M).
The mIoU of SG-One-separate obtains 44.8\%, and this result is far better than the 41.1\% of co-FCN.
This comparison shows that the approach we applied for incorporating the guidance information from support image pairs is superior to OSLSM and co-FCN in segmenting unseen classes.
Surprisingly, the proposed unified network can even achieve higher performance of 46.3\%. 
We attribute the gain of 1.5\% to the reuse of the network in extracting support and query features. 
The reuse strategy not only reduces the demand of computational resources and decreases the risk of over-fitting, but offers the network more opportunities to see more training samples.
OSLSM requires the most parameters (272.6M), whereas it has the lowest score.
It is also worth mentioning that the number of OSLSM is from the officially released source code, while the number of co-FCN is based on our reimplemented version. Both of the two baseline methods do not share parameters in processing query and support images.
\begin{table}\setlength{\tabcolsep}{20pt}
  \centering
  \caption{Comparison in the number of network parameters and one-shot segmentation Mean IoUs.}\label{tab_num_para}
  \begin{tabular}{l|c|c}
    \hline
    \hline
    \textbf{Methods} & \textbf{Parameters} & \textbf{Mean} \\
    \hline
    OSLSM~\cite{shaban2017one} & 272.6M  & 40.8 \\
    co-FCN~\cite{rakelly2018conditional} & 34.2M  & 41.1 \\
    SG-One-separate &  36.1M &  44.8 \\
    SG-One-unified &  \textbf{19.0M} & \textbf{46.3} \\
    \hline
    \hline
  \end{tabular}
  \vspace{-10pt}
\end{table}

\begin{figure}
  \centering
  \includegraphics[width=0.45\textwidth]{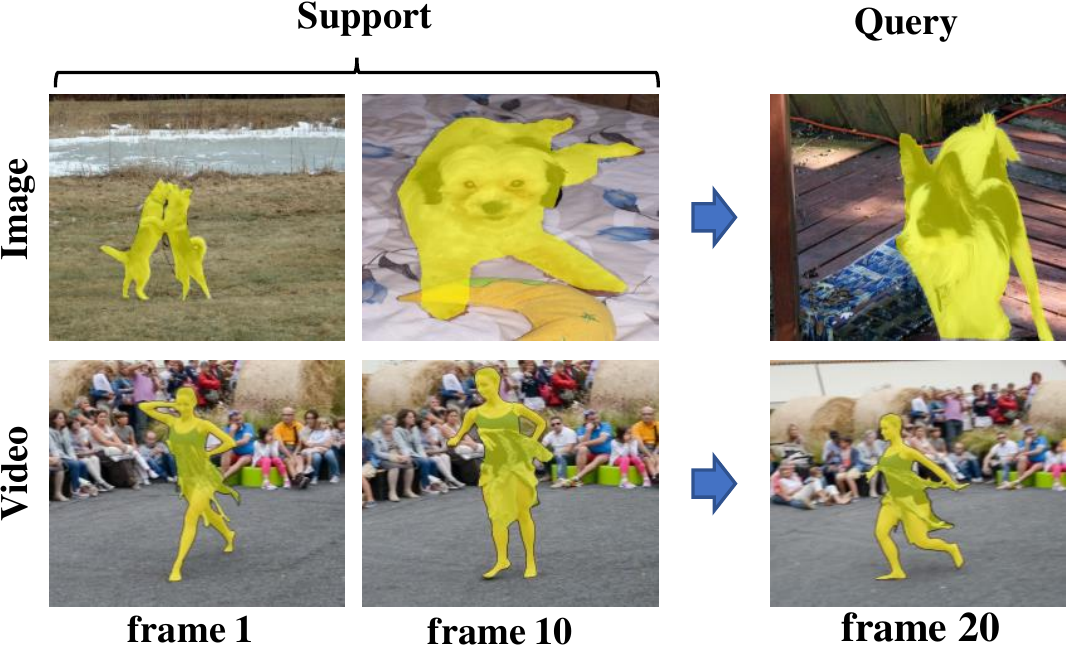}
  \caption{Comparison between the few-shot image segmentation and few-shot video segmentation tasks. The object and background environment keep consistent in between video frames, while both objects and environment are greatly various in the image segmentation task.}\label{task_diff}
  \vspace{-10pt}
\end{figure}

\subsection{Relationship with Video Object Segmentation}
One-shot video segmentation is to segment specified objects in video clips with only the first frame densely annotated~\cite{davis2017}.
Similar to our one-shot image semantic segmentation problem, the testing categories of the video segmentation problem are inconsistent with the trianing categories.
So, for the both tasks, the underline mission is to learn the relationship between the feature embeddings of the support and query images.
Siamese network~\cite{koch2015siamese} is designed to learn the relationship of such a pair of support and query images by applying two parallel networks for extracting high-level abstract features of them, separately.
Both the proposed method and a wealth of the video segmentation methods~\cite{hu2018videomatch, shin2017pixel,oh2018fast} are derivatives of the Siamese network~\cite{koch2015siamese}.

However, the key difference of the two problems is the source of information between support and query. 
First, in the video task, the contents of the target objects and background remain consistent in one sequence.
For example, given a video clip of a girl dancing on the grass land, the foreground target (the girl) and the background environment (the grass land) do not change very seriously between different frames.
In contrast, one-shot image semantic segmentation task does not exist sequential cues in targeting objects nor the background environments. 
The objects and background in query images are seriously different from the support images . 
For instance, for our one-shot image segmentation task, we may be required to segment an old man standing on grass with the reference to a little girl lying in bed, as they both belong to the same category, namely, \textit{person}.
Second, benefiting from sequential cues in videos,  the video segmentation methods~\cite{hu2018videomatch} can calculate the frame-to-frame similarities from successive frames and to boost the performance by online updating.
Figure~\ref{task_diff} illustrates the differences between the two tasks.
In the video segmentation task, the target objects and the background environment maintain consistent in the whole video clip.
In contrast, the objects and environments are totally different between the support images and the query image in our image segementation task. 
Compared to our task, none of the background nor successive information can be applied.

We apply our SG-One network to the one-shot video segmentation task on DAVIS2016~\cite{davis2016}.
We try our best seeking the fair comparison results in video segmentation papers, which do not apply background similarities nor successive object cues between frames.
In Table~\ref{davis_comp}, the results of the baseline models, OSVOS~\cite{caelles2017one}, VideoMatch~\cite{hu2018videomatch} and RGMP~\cite{oh2018fast} are obtained by excluding background features and successive frame-to-frame consistencies.
These models are merely trained on the training set of DAVIS2016 by randomly selecting image pairs, excluding the finetuning step on testing set and any sequential cues between frames.
In Tab.~\ref{davis_comp}, we compare the mIoU of these different algorithms on DAVIS2016 testing set, and our SG-One achieves the best accuracy of 57.3\%, surpassing the baseline methods.
The proposed model is more robust and better in segmenting the given query images with the reference to only one annotated image.

\begin{table}\setlength{\tabcolsep}{10pt}
  \centering
  \caption{mIoU on DAVIS2016 validation set. Sequential cues and online finetuning are disabled for a fair comparison, as the sequential cues are unavailable in our task .}\label{davis_comp}
  \begin{tabular}{c|c|c|c}
    \hline
    \hline
     \textbf{OSVOS}~\cite{caelles2017one} & \textbf{VideoMatch}~\cite{hu2018videomatch} & \textbf{RGMP}~\cite{oh2018fast} &  \textbf{SG-One} \\
    \hline
      52.5 & 52.7 & 55.0  & \textbf{57.3} \\
    \hline
    \hline
  \end{tabular}
  \vspace{-10pt}
\end{table}

\vspace{-10pt}
\section{Conclusion and Future Work}
We present that SG-One can effectively segment semantic pixels of unseen categories using only one annotated example.
We abandon the previous strategy~\cite{shaban2017one,rakelly2018conditional} and propose the masked average pooling approach to extract more robust object-related representative features. 
Extensive experiments show the masked average pooling approach is more convenient and capable of incorporating contextual information to learn better representative vectors.
We also reduce the risks of the overfitting problem by avoiding the utilization of extra parameters through a unified network. 
We propose that a well-trained network on images of a single class can be directly applied to segment multi-class images.
We present a pure end-to-end network, which does not require any pre-processing or post-processing steps. 
More importantly, SG-One boosts the performance of one-shot semantic segmentation and surpasses the baseline methods.
At the end, we analyze the relationship between one-shot video segmentation and our one-shot image semantic segmentation problems.
The experiments show the superiority of the proposed SG-One in terms of segmenting video objects under fair comparison conditions.
Code has been made available. We hope our simple and effective SG-One can serve as a solid baseline and help ease future research on one/few shot segmentation.
There are still two problems and we will try to solve them in the future. First, due to the challenging settings of the one-shot segmentation problem, the latent distributions between the training classes and testing classes do not align, which prevents us from obtaining better features for input images. Second, the predicted masks sometimes can only cover part of the target regions and may include some background noises if the target object is too similar to the background.

\section*{Acknowledgment}
This work is supported by ARC DECRA DE190101315 and ARC DP200100938.

\ifCLASSOPTIONcaptionsoff
  \newpage
\fi



\bibliographystyle{IEEEtran}
\bibliography{IEEEabrv,ref}

\vspace{-40pt}
\begin{IEEEbiography}[{\includegraphics[width=1in,height=1.25in,clip,keepaspectratio]{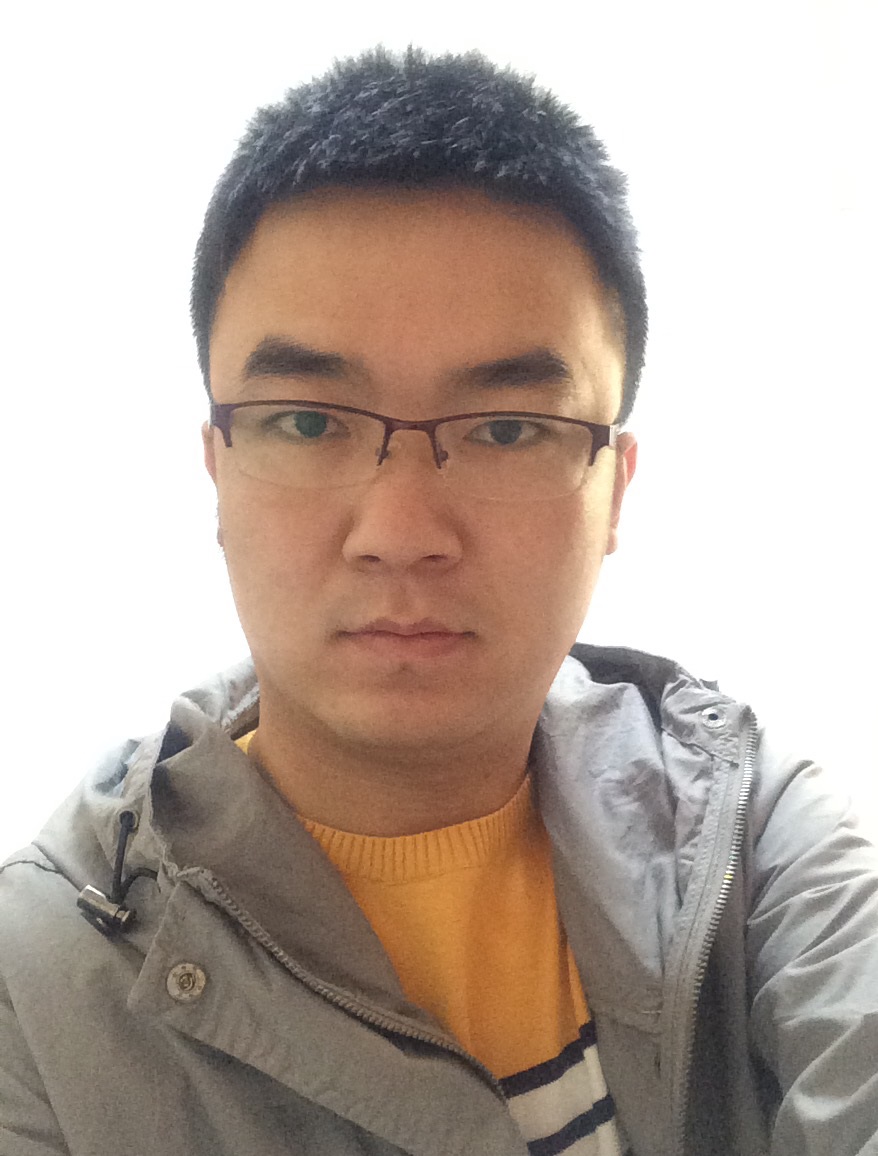}}]{Xiaolin Zhang}
is a Ph.D. student at the University of Technology Sydney, Australia. 
He received the B.S. degree and M.S. degree from the Lanzhou University, China, in 2013 and 2016, respectively.
He focuses on weakly object localization and segmentation, one-shot image semantic segmentation.
\end{IEEEbiography}
\vspace{-40pt}

\begin{IEEEbiography}[{\includegraphics[width=1in,height=1.25in,clip,keepaspectratio]{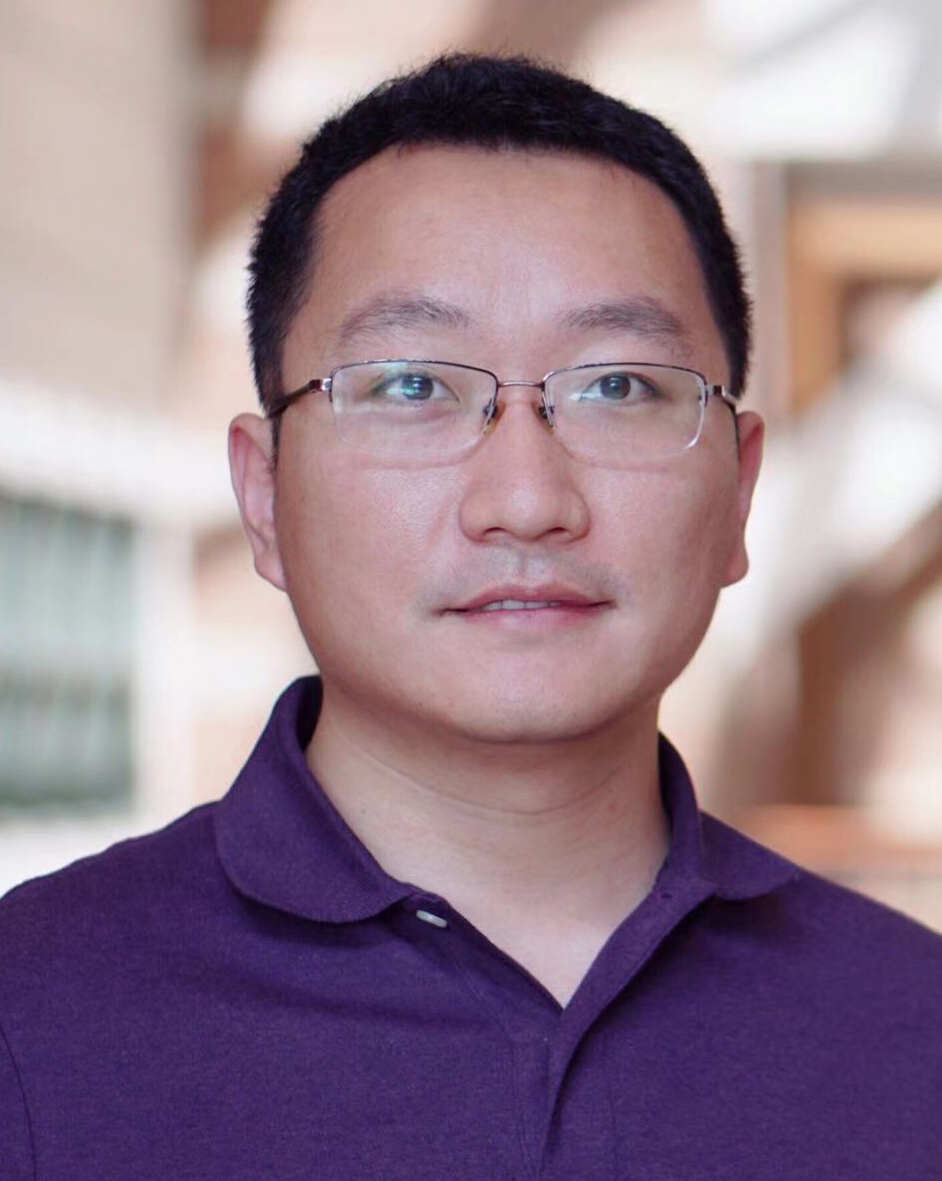}}]
{Yunchao Wei}
is currently an Assistant Professor with the University of Technology Sydney. He received his Ph.D. degree from Beijing Jiaotong University, Beijing, China, in 2016. He was a Postdoctoral Researcher at Beckman Institute, UIUC, from 2017 to 2019. He is ARC Discovery Early Career Researcher Award Fellow from 2019 to 2021. His current research interests include computer vision and machine learning.
\end{IEEEbiography}
\vspace{-40pt}

\begin{IEEEbiography}[{\includegraphics[width=1in,height=1.25in,clip,keepaspectratio]{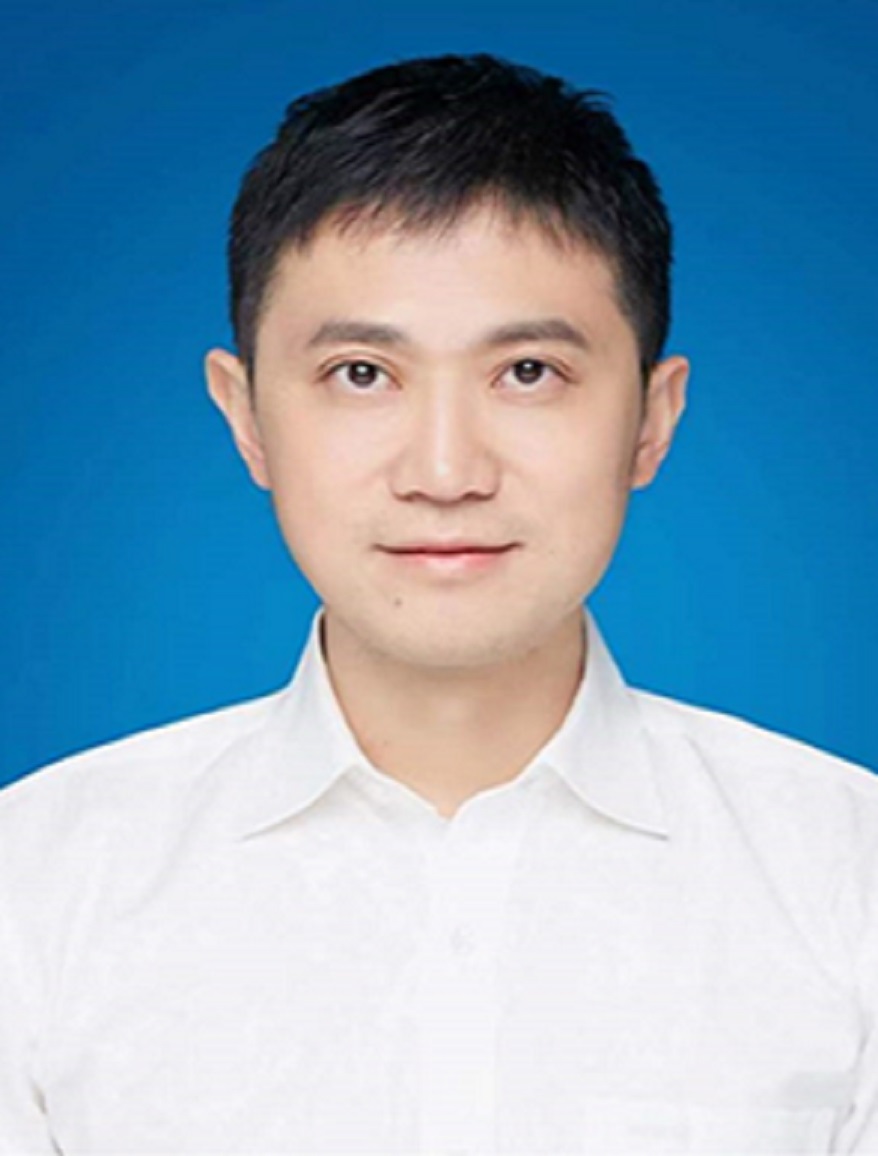}}]{Yi Yang}
received the Ph.D. degree in computer science from Zhejiang University, Hangzhou, China, in 2010. He is currently a professor with University of Technology Sydney, Australia. He was a Post-Doctoral Research with the School of Computer Science, Carnegie Mellon University, Pittsburgh, PA, USA. His current research interest include machine learning and its applications to multimedia content analysis and computer vision, such as multimedia indexing and retrieval, surveillance video analysis and video semantics understanding.
\end{IEEEbiography}
\vspace{-40pt}

\begin{IEEEbiography}[{\includegraphics[width=1in,height=1.25in,clip,keepaspectratio]{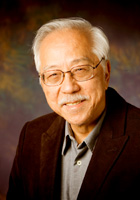}}]
{Thomas S. Huang}
received Sc.D. from the Massachusetts Institute of Technology in 1963. He was a tenured faculty member at MIT from 1963 to 1973. He was director of the Information and Signal Processing Laboratory at Purdue University from 1973 to 1980. He is currently Research Professor and Swanlund Endowed Chair Professor Emeritas with the University of Illinois at Urbana-Champaign, and Director of Image Formation and Processing (IFP) group at Beckman Institute. He has co-authored over 20 books and over 1300 papers. He is a leading researcher in computer vision, image processing and multi-modal signal processing. He is a member of the National Academy of Engineering. He is a fellow of IEEE, IAPR, and OSA. He was a recipient of the IS\&T and SPIE Imaging Scientist of the Year Award in 2006. He has received numerous awards, including the IEEE Jack Kilby Signal Processing Medal, the King-Sun Fu Prize of the IAPR, and the Azriel Rosenfeld Life Time Achievement Award at ICCV. 
\end{IEEEbiography}

\end{document}